\documentclass[final]{cvpr}

\title{Pareto-Optimal Quantized ResNet Is Mostly 4-bit}

\usepackage{amsmath}
\usepackage{times}
\usepackage{graphicx}
\usepackage{xcolor}
\usepackage{url}
\usepackage[pagebackref=true,breaklinks=true,colorlinks,bookmarks=false]{hyperref}

\def\cvprPaperID{26} %
\def\confYear{CVPR 2021}
\begin{document}

\author{AmirAli Abdolrashidi\thanks{Equal Contribution.}\hspace{4.5pt}$^a$, Lisa Wang\textsuperscript{\fnsymbol{footnote}}$^b$, \\
 Shivani Agrawal$^b$, Jonathan Malmaud$^b$, Oleg Rybakov$^b$, Chas Leichner$^b$, Lukasz Lew$^b$ \\
$^a$University of California, Riverside, CA, USA \\
$^b$Google Research, Mountain View, CA, USA \\
{\tt\small amirali.abdolrashidi@email.ucr.edu,} \\
{\tt\small \{wanglisa, shivaniagrawal, malmaud, rybakov, cleichner, lew\}@google.com\thanks{Correspondence to: wanglisa@google.com, lew@google.com.} }
}

\maketitle

\begin{abstract}

Quantization has become a popular technique to compress neural networks and reduce compute cost, but most prior work focuses on studying quantization without changing the network size. Many real-world applications of neural networks have compute cost and memory budgets, which can be traded off with model quality by changing the number of parameters. In this work, we use ResNet as a case study to systematically investigate the effects of quantization on inference compute cost-quality tradeoff curves. Our results suggest that for each bfloat16 ResNet model, there are quantized models with lower cost and higher accuracy; in other words, the bfloat16 compute cost-quality tradeoff curve is Pareto-dominated by the 4-bit and 8-bit curves, with models primarily quantized to 4-bit yielding the best Pareto curve. Furthermore, we achieve state-of-the-art results on ImageNet for 4-bit ResNet-50 with quantization-aware training, obtaining a top-1 eval accuracy of 77.09\%. We demonstrate the regularizing effect of quantization by measuring the generalization gap. The quantization method we used is optimized for practicality: It requires little tuning and is designed with hardware capabilities in mind. Our work motivates further research into optimal numeric formats for quantization, as well as the development of machine learning accelerators supporting these formats. As part of this work, we contribute a quantization library written in JAX, which is open-sourced at \url{https://github.com/google-research/google-research/tree/master/aqt}.
\end{abstract}

\section{Introduction}
\label{sec:intro}
\begin{figure}[htb!]
\centering
\includegraphics[width=1.01\linewidth]{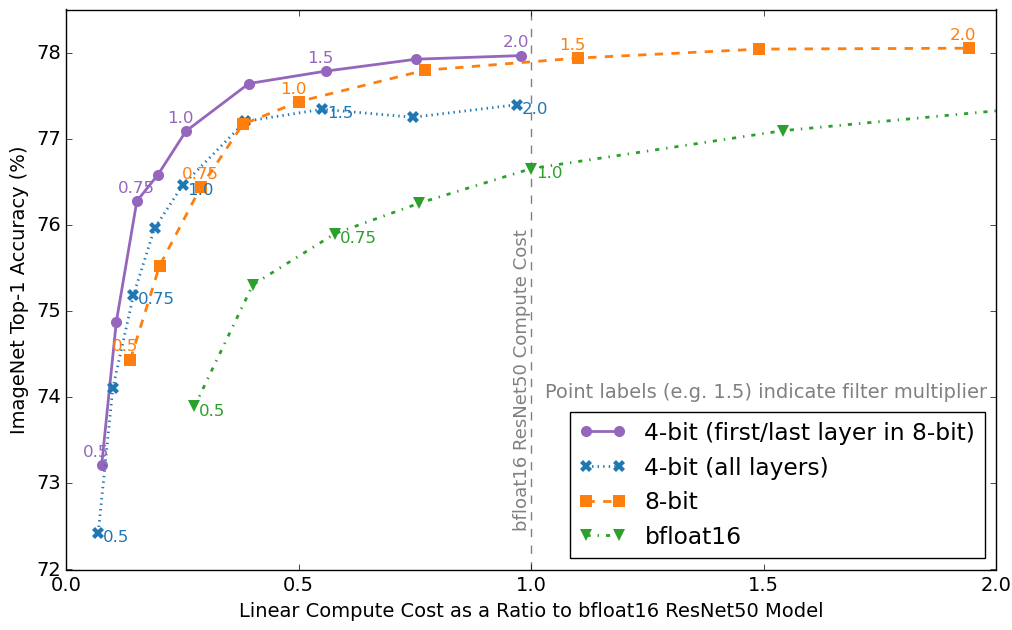}
\caption{Compute cost-accuracy tradeoff curves, using a \textit{linear} compute cost model. }
\label{fig:cost_acc_bits}
\end{figure}

While neural networks have brought tremendous progress to the computer vision field over the last decade, they are also often resource-intensive to train and serve.
In many real-world computer vision applications, it is therefore of interest to minimize the compute cost (including power consumption/CO$_{2}$ output), and the memory footprint, ideally without compromising model quality. 

Quantization \cite{han2016deep} has become a popular technique to make neural networks, including computer vision models, more efficient. By reducing the number of bits used, quantization helps to compress the model, thus reducing its memory footprint. With support for quantized ops in hardware, e.g. NVIDIA A100 GPUs \cite{A100_whitepaper}, quantization can also significantly speed up computation and lower power consumption, which can reduce overall compute costs. 
Quantization directly reduces per-op power-usage, an important factor in both chip design and the cost of ownership.
Table \ref{tab:a100_perf} shows NVIDIA A100 peak performance in the context of quantization. 

\begin{table}[htb!]
    \centering
    \begin{tabular}{|c|c|}
    \hline \textit{Input types}  & \textit{TOPS}  \\
    \hline 
    \hline \text{float16}   & 312 \\
    \hline \text{bfloat16}  & 312 \\
    \hline \text{int8}      & 624 \\
    \hline \text{int4}      & 1248 \\
    \hline \text{binary}    & 4992 \\
    \hline
\end{tabular}
\vspace{4mm}
\caption{NVIDIA A100 performance on quantized types.} \label{tab:a100_perf}
\end{table}

\begin{figure*}[htb!]
\centering
\includegraphics[width=0.9\linewidth]{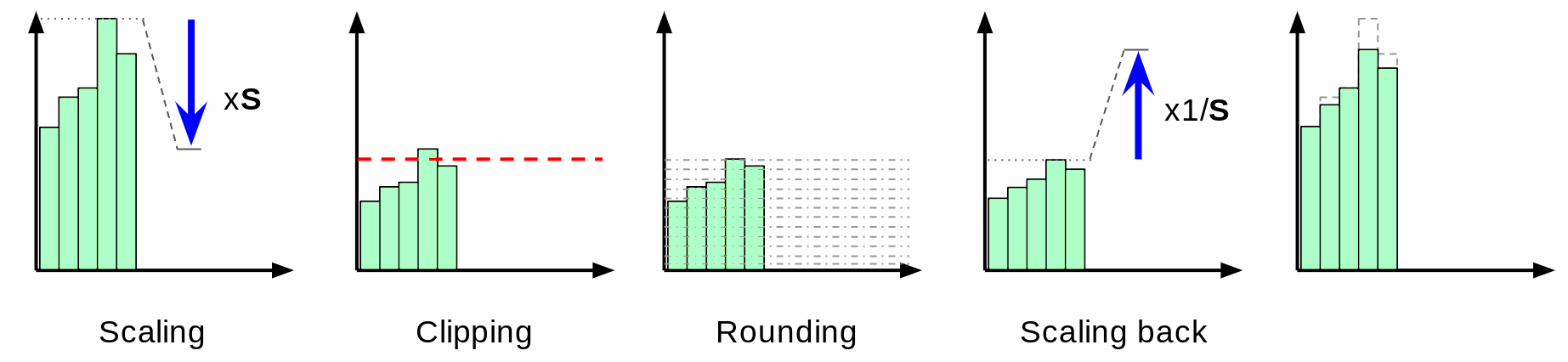}
\caption{Different steps of quantization for a single data value (left to right), including the difference between the original data and the final quantized output (right).}
\label{fig:quant_example}
\end{figure*}

Despite its advantages, there could also be downsides to quantizing a model, e.g., a possible reduction in model quality. 
Figure \ref{fig:quant_example} displays an example showing quantization steps performed on a series of data, e.g. model weights. During quantization, the tensors would be scaled down, clipped, rounded to their closest quantization level and scaled back to their original domain. As it can also be seen in the figure, some of the values after quantization could be considerably far from their original values.  Depending on the model, this could result in a quality reduction and, in some cases, prevent the model from converging.

A major hurdle to wider adoption of quantization both in industry and in research
is the added engineering complexity introduced by quantization, e.g. some quantization techniques add hyperparameters for clipping bounds, which would require tuning as well. This motivates us to focus on approaches that offer clear benefits while minimizing the amount of added complexity.  

In this work, using ResNet \cite{ResNet} as an example, we seek to understand how different quantization precisions affect the compute cost-accuracy tradeoff curves, and find a simple strategy to compress models at different compute cost and quality requirements.

After running experiments on ResNet using different precisions and numbers of parameters, we determined that 4-bit and 8-bit models strongly Pareto-dominate bfloat16 models, and mostly-4-bit models outperform 8-bit models. We present our results using two compute cost models (\textit{linear} and \textit{quadratic}), based on different assumptions about the compute speedups when the number of bits are reduced. Both cost models will be defined and justified in Section \ref{sec:cost}. While no hardware with the \textit{quadratic} cost model is available yet to our knowledge, our results provide strong motivation for the development of such hardware.
\section{Background}
\label{sec:resnet}

\subsection{Related Work}
\label{sec:related}
One of the advantages model quantization provides is enabling more embedded systems to use such models \cite{howard2017mobilenets}. However, it also provides server-side benefits by reducing data transfers and computation complexities \cite{li2018auto}. Therefore, there have been many works in recent literature on improving the quality of various quantized neural networks, leveraging methods such as improved quantization techniques, compact model design, or even a combination of methods \cite{kim2020spatial}.
Leveraging embedded systems and supporting hardware for operations such as convolution has also been proposed to enable 4-bit quantization with low quality loss and reduce power consumption \cite{Xilinx_whitepaper}. 
MobileNets \cite{howard2017mobilenets} reduce convolution complexity by breaking it into two simpler layers, and use two multipliers in the model to reduce size in exchange for an acceptable accuracy loss.
The authors in \cite{wu2020integer} study a multitude of pre-trained neural networks, and observe that they can all be quantized to int8 with their accuracies remaining within 1\% of the baseline model.
In order to find the best quantization parameters for a given network, neural architecture search (NAS) could be used \cite{zoph2016neural,elsken2019neural}.
EfficientNet \cite{tan2019efficientnet} uses NAS to explore combinations of changing number of channels, resolution, and depths of different convolutional neural networks, such as ResNet \cite{ResNet}, Inception \cite{InceptionV4} and AmoebaNet \cite{AmoebaNetA}, in order to find the optimal point with resource constraints in mind, e.g. mobile applications. When compared to a network with similar accuracy, EfficientNets have shown to be up to 8.4x smaller and 6.1x faster on hardware during inference.
However, NAS can be very complex and resource-consuming, prompting us to look for a simpler solution.

Post-training quantization (PTQ) enables the user to convert an already trained float model and quantize it without retraining \cite{fang2020post,oh2020weight,dai2021vs,garg2021confounding}. However, it can also result in drastic reduction in model quality. To address the quality degradation, quantization-aware training (QAT) has been proposed and applied in several papers \cite{choi2018pact, dong2019hawq, wu2020integer} and was also our method of choice.

Obtaining optimal clipping bounds for quantizing activations has been studied in several works. Choi et al. \cite{choi2018pact} propose a quantization scheme in which an activation clipping parameter for controlling bounds is introduced and optimized during training. With this technique, they are able to quantize weights and activations to 4-bits with little loss in quality.

\subsection{ResNet50}

\begin{figure}[htb!]
\centering
\includegraphics[width=0.99\linewidth]{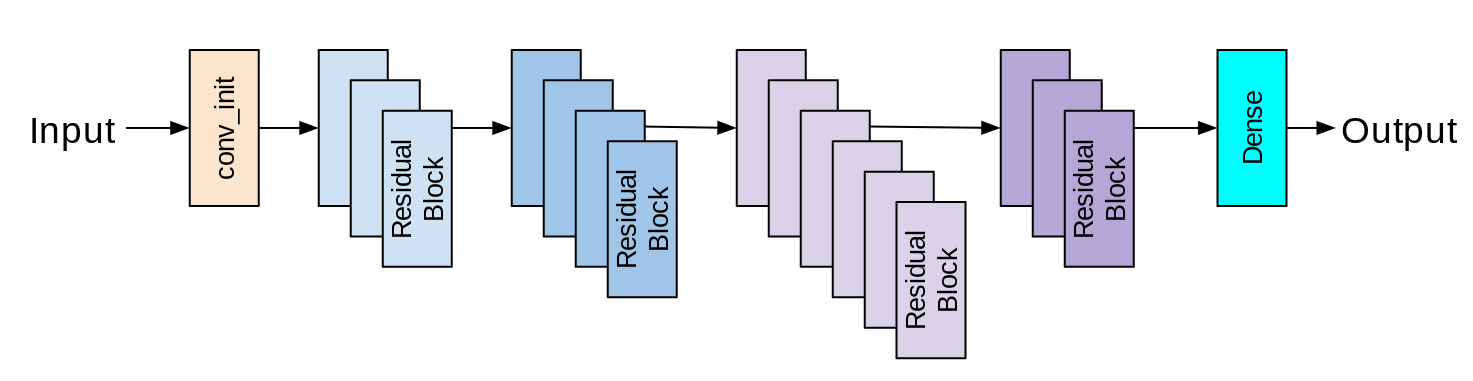}
\caption{ResNet50 architecture}
\label{fig:resnet50}
\end{figure}

\begin{figure}[htb!]
\centering
\includegraphics[width=0.6\linewidth]{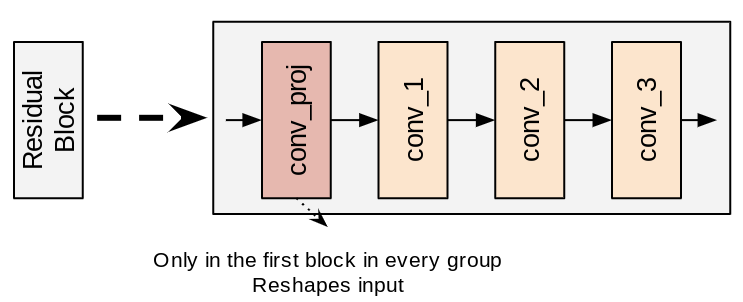}
\caption{ResNet block}
\label{fig:resnet_block}
\end{figure}

We use the bfloat16 ResNet50 v1.5 model as our baseline and implemented our quantized model on top of JAX \cite{jax2018github} MLPerf ResNet50 submission. Figure \ref{fig:resnet50} shows the ResNet50 architecture used in this work. It consists of an initial convolutional layer (\textit{conv\_init}), 16 residual blocks and a dense layer at the end. Within each residual block, as shown in Figure \ref{fig:resnet_block}, there are three convolutional layers in series. In addition, there is also a projection layer at the beginning of each block group only (thus 4 in total in ResNet50), which is responsible for reshaping the inputs so they can be used by the rest of the convolutional layers in the block group.

To change the number of model parameters, we multiply the number of convolutional filters in each layer by a global filter multiplier $c\in(0.5, 2)$.
\section{Quantization Details}
In this work, our goal is to study the impact of quantization on cost-quality tradeoffs at different precisions and find practical, hardware-friendly approaches for model compression, so we chose to build our work on top of popular quantization methods found in literature. We apply quantization to both weights and activations, and introduce quantization during training. We decided to focus on 8-bit and 4-bit quantization, since these formats are already supported in existing hardware.
\label{sec:quant}
\subsection{Uniform Quantization}
For our experiments, we employ uniform integer quantization, where all quantization buckets are of equal size. This is a necessary choice if one wants to benefit from acceleration in existing hardware. We use scaling and clipping to reduce the impact of outliers. This form of quantization is most popular; for instance, it is found in prior work \cite{han2016deep, wu2020integer, Xilinx_whitepaper} and open-source machine learning frameworks including TensorFlow \cite{abadi2016tensorflow} and PyTorch \cite{Pytorch}.

Uniform quantization consists of the following steps, which are also shown in Figure \ref{fig:quant_example}:
\begin{enumerate}
    \item \textit{Scale}: Scale floating point input $x$ by $S$, usually to ensure efficient use of the target range, e.g. $[-127, 127]$ for 8-bit signed integers. We compute scales per-channel for both weights and activations. The process of obtaining scales will be explained in more detail in the next section on calibration. Note that we do not apply shift, which helps us preserve the 0 value.
    \item \textit{Clipping}: An ideal scale would allow us to express all the quantized values in our quantization range, and significantly different values once scaled would be fall into separate quantization buckets.
    However, that may not always be possible, especially when outliers are present. Scale must be chosen in such a way to balance outlier clipping and bucket resolution. We use unsigned int types for activations coming out of ReLu as they are always positive. This effectively gains 1 bit of precision. 
    \item \textit{Rounding}: After all the values are within the same scale, each value is rounded to the nearest quantization step, resulting in quantization error. If we need $B$ bits to represent a quantized value, there will be $2^{B}$ steps in the range. However, based on our preferences, the range limits could be either \textit{positive} or \textit{symmetric}, giving us a integer range of $[0, 2^{B}-1]$ or $[-2^{B-1}+1, 2^{B-1}-1]$ respectively.
    \item \textit{Scaling back}: After rounding, the values are scaled back to the original range by multiplying with $\frac{1}{S}$. The difference between the original unquantized and the values after quantization is also known as \textit{quantization error}, which is introduced by the clipping and rounding steps.
\end{enumerate}

\subsection{Calibration of Clipping Bounds}
There are several methods to choose the clipping bounds for the scaling step. One could use hard-coded fixed bounds, e.g. clip values to a range of $[0, 6]$, which is inspired by Relu6 \cite{Krizhevsky10convolutionaldeep}. These fixed bounds can also be picked via hyperparameter tuning, but the number of hyperparameter quickly explodes e.g. if we want to choose different scales for different layers or even different channels within a layer. 
Prior work \cite{wu2020integer} as well as our own experiments have shown that automatically picking bounds during training, a process also known as calibration, yields better accuracy and requires far less hyperparameter tuning compared to fixed bounds. It also allows us to easily customize bounds for each layer and even channel within a layer. For example, if the weight tensor has shape $[3 \times 3 \times 6]$ where $6$ is the number of output channels (or convolutional filters), we would have $6$ different scales for each of the channels.

To obtain the clipping bounds for activations, we use the following calibration method:
\begin{enumerate}
    \item For the first $N$ training steps, we do not quantize activations, but compute per-channel statistics of $max(abs(x))$ values and keep track of their exponential moving averages (EMA). We also experimented with other calibration methods, e.g. $c \times stddev(x)$, but found that EMA of $max(abs(x))$ worked best for the ResNet case, and doesn't add sensitive hyperparameters. %
    \item At training step $N$, we finalize the calibration of clipping bounds by setting them to their most recent moving averages, turn on activation quantization with the new clipping bounds, and continue training with quantization. Importantly, we only calibrate activation clipping bounds once during training, as we found that more frequent calibrations can lead to feedback loops (e.g. vanishing or exploding bounds) as well as worse results. This choice makes the model insensitive to EMA hyperparameter. The specific step $N$ at which to finalize the calibration and turn on activation quantization is a hyperparameter, but we found that the model is insensitive to the specific choice of the step.  We recommend turning on activation quantization and setting the bounds between 10\% and 40\% of the total number of training steps it takes to reach convergence. We used 20\% in all our experiments.
\end{enumerate}

To obtain the clipping bounds for weights, we use the maximum absolute value of the current weight tensor per channel as clipping bounds. As this simple dynamic method has worked well for us, we did not try more complicated calibration methods for weights.

\subsection{Quantization Library in JAX and Flax}
We implemented a collection of quantization techniques and quantized neural networks on top of the JAX framework \cite{jax2018github} and the Flax library \cite{flax2020github} to enable fast experimentation, and used it to run the experiments in this paper. All code is at \url{https://github.com/google-research/google-research/tree/master/aqt}. To our knowledge, this is the first open source quantization library for JAX and Flax. \\

\textbf{Highlighted Features:}
\begin{enumerate}
    \item \textit{Quantized JAX dot and Flax layers:} We wrote a custom version of \texttt{jax.lax.dot} (matrix multiply) with optional weight and activation quantization. We also implemented quantized versions of common Flax layers, including \textit{Dense} and \textit{Conv}, which can be used as drop-in replacements.
    \item \textit{Multiple quantization strategies:} The user can choose between several algorithms to calibrate bounds automatically, including running mean of maximum values or statistics-based methods e.g. standard deviation or absolute-deviation of activations.
    \item \textit{Flexible configuration system:} Our configuration system enables fine-grained control of quantization settings, e.g., allowing the user to set separate precisions and quantization strategies for different layers which can useful for experimenting with mixed-precision models. Weights and activations within a layer can be set to different precisions as well.
    \item \textit{Support for unsigned and signed quantization:} We allow the user to configure whether to use unsigned or signed precision. This is especially useful for 4-bit quantization of positive activations (e.g. after a ReLU activation function) without shifting, since the unsigned integer doubles the resolution in the positive range.
    \item \textit{What you train is what you serve:} Optionally, the user can choose to apply our Accurate Quantized Training (AQT) method instead of using fake-quantization \cite{jacob2018quantization}. AQT ensures that the forward-pass during training is the same as the forward pass during inference, i.e. the matrix multiplies are in true integer domain. This provides better quality guarantees, enables training-time cost savings and simplifies compiler logic for inference, since no conversion is needed from training to inference graphs.
\end{enumerate}

\section{Cost Models for Quantized Neural Networks}
\label{sec:cost}

\subsection{Linear Cost Model for Existing Hardware}
\label{sec:linear_cost}
Some processors \cite{A100_whitepaper,abadi2016tensorflow} can support operations with reduced-precision operands. These operations are faster and consume less power than full-precision operations. In particular, integer operations can provide a higher throughput and lower cost than floating point operations \cite{wu2020integer}. The NVIDIA A100 \cite{A100_whitepaper} processor architecture supports 16-bit, 8-bit, and 4-bit multiplications.
As Table \ref{tab:a100_perf} shows, the hardware can process convolution and matrix multiplication operations twice as quickly when their inputs are quantized to 8 bits as when they are quantized to 16 bits. In other words, 8-bit computations have half the \textit{compute cost} of 16-bit computations. Moving from 8 bits to 4 bits provides a similar benefit. This means that there is a linear relationship between the (power of 2) total number of bits in the operation and the cost of that operation for these processors.

\subsection{Energy-Motivated Quadratic Cost Model}
\label{sec:quadratic_cost}

In the process of quantizing a model, each halving of number of bits used leads to power usage being at least halved for all hardware aspects of running that model (SRAM and DRAM memory reads and writes, data movement around the chip, arithmetic operations and nonlinear functions). For the multiplications, however, power usage scales better than linear as the precision is reduced. Given the extreme importance of multiplier power to ML accelerators (e.g. in matrix multiplications, einsums, and convolutions), this motivates a closer look at the fundamental cost model for multiplications.

Hardware multiplication of two $n$-bit numbers requires reducing $n^2$ bits to $2n$ bits i.e., $$a\cdot b = \sum\limits_{0 \leq i,j < n} 2^{i+j} \text{~AND~}(a_i, b_i)$$
This is done using appropriately wired adder circuits, e.g. Wallace trees. Each of them takes 3 bits of input and produces two bits of output. This means that an $n$-bit multiplier requires roughly $n^2$ adders and AND gates. Multiplying two $2n$-bit numbers therefore requires $4n^2$ adders. Further, one $2n$-bit multiplier is roughly equivalent to \emph{four} $n$-bit multipliers. This relationship motivates a quadratic cost model. The quadratic relationship holds for both the area and the power of integer multiplier circuits.

\subsection{Cost Modeling of Layers and Full Network}
The compute cost of the whole neural network is modeled as the sum of the costs of each layer. Since the cost of multiplication operations usually dominates the cost of arithmetic operations \cite{agarwal1995three}, layer cost is approximated by the sum of the costs of all the multiplication operations. While we use this approach to assign a cost to each layer, a ResNet model mostly consists of convolution layers and dense layers. The computational costs for these layers is a follows:

$$\text{Cost}_\text{Conv2D} = B K_{\text{h}} K_{\text{w}} A_{\text{w}} A_{\text{h}} C_{\text{in}} C_{\text{out}} M $$ 

$$\text{Cost}_\text{Dense} = B C_{\text{in}} C_{\text{out}} M $$

$B$ is batch size, 
$K_{\text{h}}$ and
$K_{\text{w}}$ are kernel width and height,
$A_{\text{w}}$ and
$A_{\text{h}}$ are layer's output image width and height,
$C_{\text{in}}$ and
$C_{\text{out}}$ are number of input and output channels,
$M$ is either a linear or quadratic cost coefficient defined above. 
We report only relative costs between quantization levels so
for bfloat16, 8-bit integer, and 4-bit integer layers, $M$ is respectively 16, 8, 4 in the linear cost model and 16, 4, 1 in the quadratic cost model.

Similarly, as a metric of memory usage, \textit{memory cost} can also be approximated to the total number of weight bits (ignoring biases and other small contributions) in the entire model. The number of bits for each convolution and dense layer will be as below:

$$\text{Mem}_\text{Conv2D} = K_{\text{h}} K_{\text{w}} C_{\text{in}} C_{\text{out}} M' $$ 

$$\text{Mem}_\text{Dense} = C_{\text{in}} C_{\text{out}} M' $$ 
where $M'$ for bfloat16, 8-bit integer, and 4-bit integer layers is 16, 8, 4 respectively in both the linear and quadratic cost models. This is an appropriate approximation of memory consumption because within a given layer, we quantize weights and activations to the same precision.
\section{Experiments and Results}
\label{sec:eval}

To evaluate the model quality, in all the experiments we computed and saved the top-1 accuracy on the ImageNet eval dataset \cite{ImageNet}. We first compare our 4-bit results on ResNet50 without changing the number of parameters to prior work. We then share and discuss the results from compute cost-accuracy tradeoff experiments, where we change the number of parameters and quantization bits.

\subsection{Comparison to Prior Work}
While the primary focus of our work is to evaluate the impact of quantization on Pareto curves, we want to provide evidence for the competitiveness of our quantization method compared to prior work.
Table \ref{tab:acc} shows the top-1 accuracy of a ResNet50 model quantized using various methods, differing by goals and constraints, which we did our best to take into account as much as possible. 
As can be seen, our quantization method achieves or beats the results found in prior work on ResNet-50 quantization. E.g. our 4-bit model (with first and last layers in 8-bit) achieves a higher accuracy compared to results in the PACT paper \cite{choi2018pact}, as well as the XILINX paper \cite{Xilinx_whitepaper}.

One may notice that even though all the results in Table \ref{tab:acc} are obtained on ResNet50, they differ significantly in Top-1 accuracy, potentially due to differences in ResNet versions and hyperparameter choices. Therefore, we focus on quantization loss, i.e. the difference between Top-1 of unquantized and quantized model, as the main metric to evaluate quality of the quantization algorithms.
In our case, both our 4-bit model (with first and last layers in 8-bit) and our fully 8-bit model outperform the bfloat16 baseline model, highlighting a regularizing effect of quantization. This regularizing effect can be clearly seen in the last three rows of Table \ref{tab:acc}. 
Differences in the generalization gap show that the unquantized model is overfitting significantly more to the training data than the quantized models do.
We would like to point out that other works may differ slightly due to quantization of  ops other than Conv2D or MatMul, BatchNorm folding, etc.. We did not attempt to catch all the differences.

\begin{table*}[!htb]
    \centering
    \begin{tabular}{|c|c||c|c||c|c|}
    \hline
    \textit{Prior work} & \textit{float (baseline)} & \textit{int8} & \textit{int8 quantization loss} & \textit{int4} & \textit{int4 quantization loss}  \\
    \hline

        \hline IAO \cite{jacob2018quantization}  & 76.40\% & {74.90\%} & -1.50\% & - & -\\
        \hline OCS \cite{zhao2019improving}  & 76.10\% & 75.70\% & -0.40\% & 66.20\%$^a$ & -9.90\%\\
        \hline {XILINX \cite{Xilinx_whitepaper} v2$^d$}  & 77.60\% & {77.47\%}  & -0.13\% & {74.12\%} & -3.48\% \\
        \hline KURE \cite{shkolnik2020robust}  & 76.30\% & - & - & 74.30\% & -2.00\%\\
        \hline {XILINX \cite{Xilinx_whitepaper} v1$^d$} & 76.15\% & {76.02\%}  & -0.13\% & {74.59\%} & -1.56\% \\
        \hline {LQ-Net \cite{zhang2018lq}} & 76.40\% & -  & - & {75.10\%} & -1.30\% \\
        \hline {SSPS \cite{sun2021effective}} & 77.15\% & -  & - & {76.22\%}$^c$ & -0.93\% \\
        \hline PACT \cite{choi2018pact}  & 76.90\% & - & - & 76.50\%$^b$ & -0.40\%\\
        \hline {HAQ \cite{wang2019haq}} & 76.15\% & -  & - & {76.14\%}$^c$ & -0.01\% \\
        \hline ZeroQ \cite{cai2020zeroq}  & 77.72\% & {77.67\%} & -0.05\% & - & -\\
        \hline TQT \cite{jain2019trained}  & 75.20\% & {75.40\%} & +0.20\% & - & -\\
        \hline FAQ \cite{mckinstry2018discovering}  & 76.15\% & 76.52\% & +0.37\% & 76.25\% & +0.10\%\\
        \hline {PQ+TS+Guided** \cite{zhuang2018towards}} & 75.60\% & -  & - & {75.90\%} & +0.30\% \\
        \hline {NICE \cite{baskin2018nice}} & 76.15\% & -  & - & {76.50\%} & +0.35\% \\
        \hline NVIDIA QAT \cite{wu2020integer} & 76.16\% & 76.85\% & +0.69\% & - & -\\
        \hline This work & 76.65\% & {77.43\%} & \textbf{+0.78\%}  & {77.09\%}$^b$ & \textbf{+0.44\%} \\
        \hline
        \hline Train log-loss (this work) & 0.633 & 0.723 & - & 0.762 & - \\
        \hline Eval log-loss (this work) & 0.974 &  0.881 & - & 0.889 & - \\
        \hline Generalization gap (this work) & 0.341 & 0.158 & - & \textbf{0.125} & - \\
    \hline
    \end{tabular}
    \vspace{2mm}

    \footnotesize{$^a$ int8 activations\;\;}
    \footnotesize{$^b$ int4 w/int8 first/last layer\;\;}
    \footnotesize{$^c$ Mixed precision\;\;} %
    \footnotesize{$^d$ ResNet v1 and v2; quantizes element-wise operations\;\;}
    \vspace{2mm}
    \caption{ImageNet Top-1 accuracy comparison of baseline and quantized ResNet-50 models sorted by int8 and int4 quantization loss which is defined as difference in Top-1 between baseline and quantized model. Last tree rows list models' train log-loss (the value optimized by the training process) and eval log-loss. Generalization gap is the difference between them.}
    \label{tab:acc}
\end{table*}

\begin{figure}[htb!]
\centering
\includegraphics[width=1.01\linewidth]{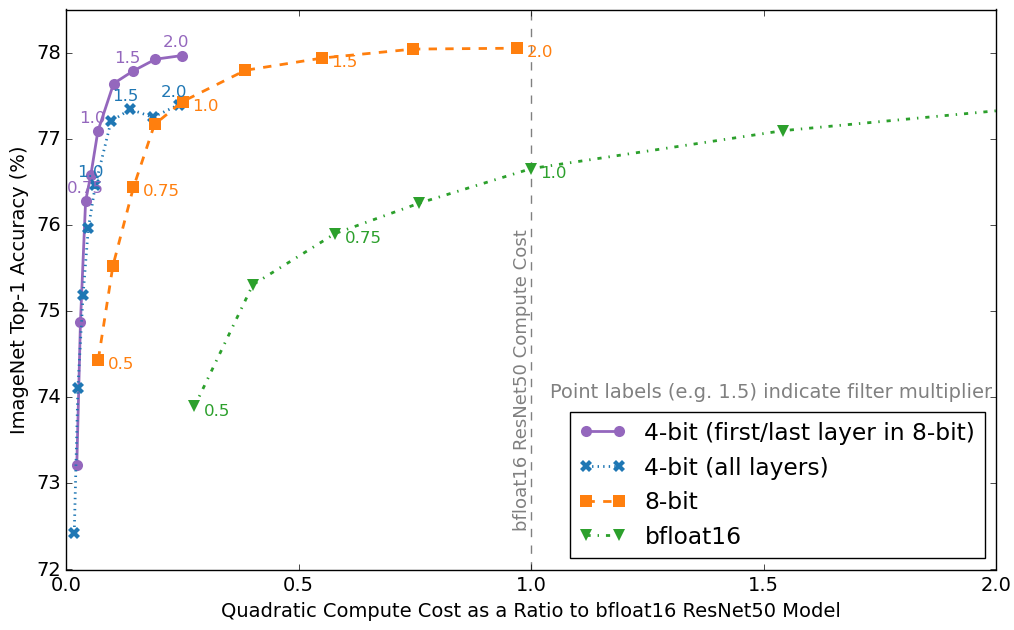}
\caption{Quadratic compute Cost-Accuracy Tradeoff Curves}
\label{fig:quad_cost_acc}
\end{figure}

\begin{figure}[htb!]
\centering
\includegraphics[width=1.01\linewidth]{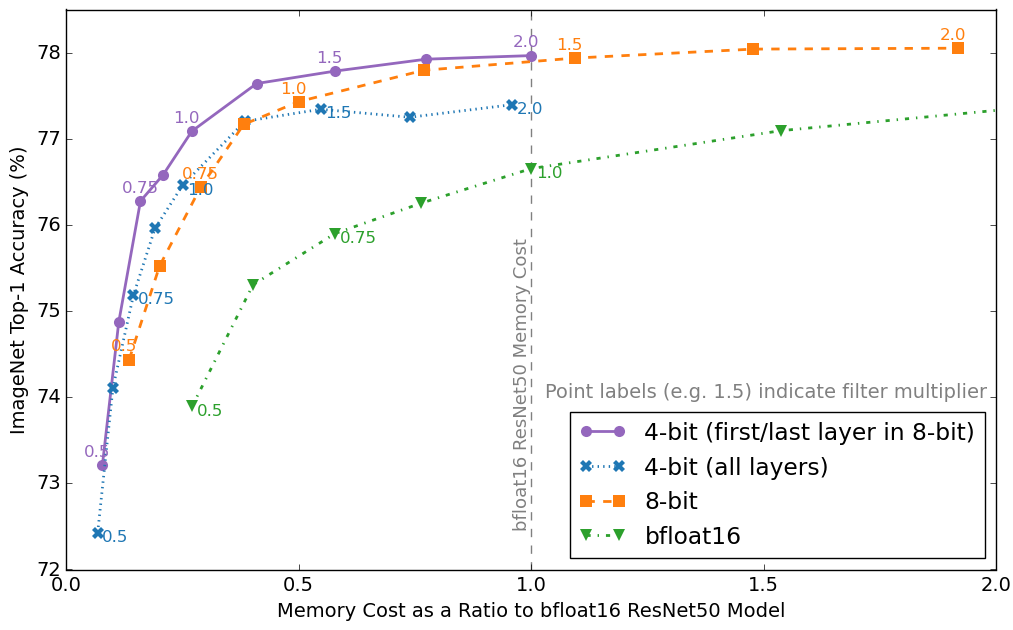}
\caption{Memory Cost-Accuracy Tradeoff Curves}
\label{fig:mem_acc_bits}
\end{figure}

\begin{figure}[htb!]
\centering
\includegraphics[width=1.01\linewidth]{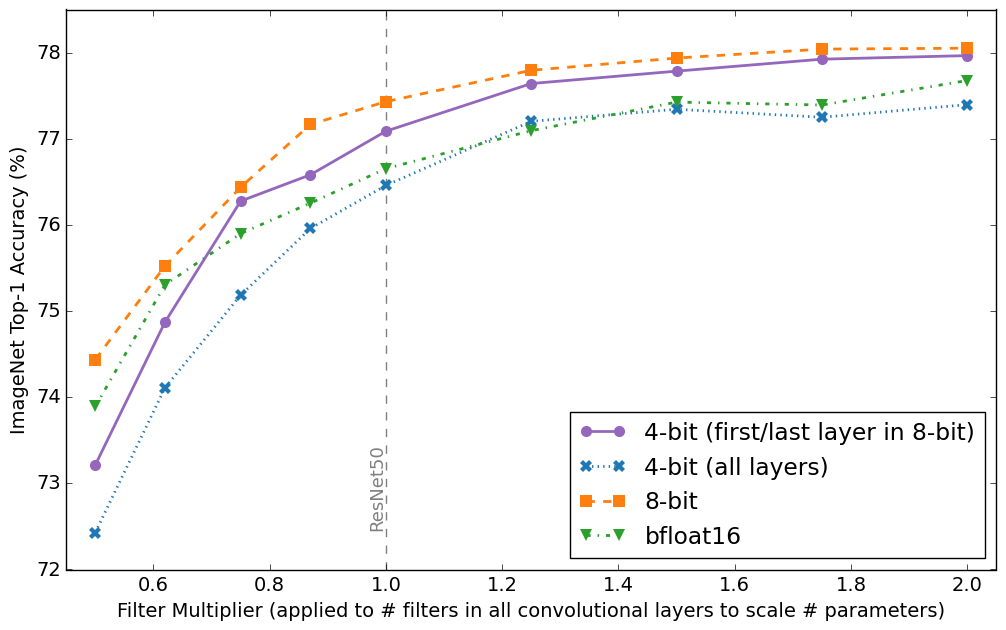}
\caption{Accuracy with respect to filter multiplier. Models with the same multiplier have the same number of parameters.}
\label{fig:multiplier_eval}
\end{figure}

\begin{figure}[htb!]
\centering
\includegraphics[width=0.6\linewidth]{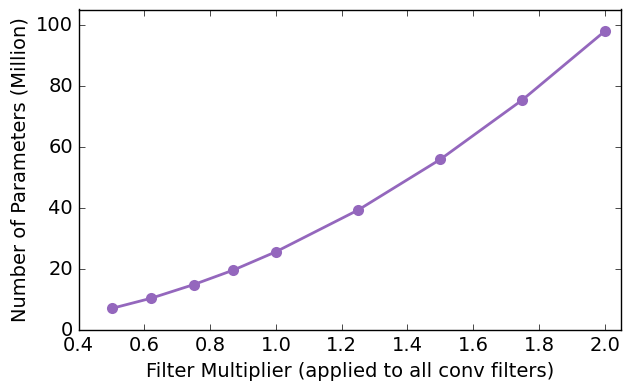}
\caption{Number of parameters with respect to filter multiplier.}
\label{fig:multiplier_params}
\end{figure}

\subsection{Compute Cost-Accuracy Tradeoff Experiments}
We ran experiments on ResNet50 with different number of parameters and quantization bits. To change the number of parameters, we multiply the number of filters in each convolutional layer by a global scalar, which we refer to as \textit{filter multiplier}. We swept over nine filter multipliers $\{0.5, 0.62, 0.75, 0.87, 1.0, 1.25, 1.5, 1.75, 2.0\}$. For instance, standard ResNet50 (multiplier $1.0$) has 25.5M parameters and a filter multiplier of $2.0$ results in 97.8M parameters. See Figure \ref{fig:multiplier_params} for a plot showing how filter multipliers affect the number of parameters.  We explored four quantization settings which are represented as four curves on the figures. All settings apply both to weights and activations.
\begin{enumerate}
\item \textit{4-bit}: All convolutional and dense layers including first and last layers (\textit{conv\_init} and \textit{dense}) in 4-bit.
\item \textit{4-bit, with first/last in 8-bit}: All layers in 4-bit, except first and last layers are in 8-bit. It is common practice to quantize these two layers to a lesser extent \cite{choi2018pact,Xilinx_whitepaper, choi2018pact}, since they tend to be most sensitive to quantization. We also found through our own layer sensitivity analysis that the first layer in ResNet is the most sensitive to quantization, followed by the last layer. Some solutions exist to address the challenge of quantizing first and last layers in neural networks, e.g. by remapping the input features \cite{zhang2021fracbnn}. We chose not to explore these special case methods, as it is not the focus of this work.
\item \textit{8-bit}: All layers quantized to 8-bit.
\item \textit{bfloat16}: Baseline, all layers in bfloat16.
\end{enumerate}

The cross-product of the filter multipliers and quantization settings results in a total of 36 experiments.

Figure \ref{fig:cost_acc_bits} showcases compute cost and Top-1 eval accuracy for ResNet50 with different parameters and quantization settings. The x-axis shows the linear compute cost (see Section \ref{sec:linear_cost}) normalized to that of the baseline bfloat16 model. In this figure, we are comparing the four quantization settings described above. Each curve corresponds to a quantization setting, the different points on the curve correspond to different filter multipliers. The Pareto curves show the tradeoff between the compute cost and the accuracy in our experiments (towards upper-left is better).

As the curves in Figure \ref{fig:cost_acc_bits} show, the 4-bit models with the first and last layers quantized to 8 bits achieve the best compute cost-accuracy Pareto curve, outperforming bfloat16, 8-bit and all-4-bit models.
The differences in the Pareto curves become more pronounced if we use the theoretical quadratic cost model (see \ref{sec:quadratic_cost} Quadratic Model), as shown in Figure \ref{fig:quad_cost_acc}. As expected, the Pareto curves are now separated more, showing an even clearer advantage of quantized models and 4-bit in particular. This motivates the development of hardware where quadratic cost savings are supported.

Figure \ref{fig:mem_acc_bits} displays the memory cost-accuracy tradeoff points for the quantized models versus the baseline, which shows a very similar pattern to the compute cost-accuracy figures.

Figure \ref{fig:multiplier_eval} summarizes the same experiments, however the x-axis is now showing the filter multiplier applied to all convolution layers. Models with the same multiplier have the same number of parameters/model architecture. We observe that the 4-bit models w/8-bit first/last achieve a better accuracy than the baseline bfloat16 starting at $multiplier=0.75$ and gets very close to the 8-bit quantized model towards $multiplier=2.0$. The 8-bit models have better accuracies than bfloat16 models for all multipliers we tried. This means that even without changing the number of parameters, quantization offers an accuracy improvement for most multipliers, most likely due to its regularization effects.

\subsection{Tradeoff-Aware Quantization Recipe}
Based on our analysis, we propose a simple recipe for model compression on ResNet with minimal hyperparameter tuning.
\begin{enumerate}
    \item Quantize all layers to 4 bits, and first and last layers (\textit{conv\_init} and \textit{dense}) to 8 bits.
    \item Change the number of parameters with a global filter multiplier to achieve the desired tradeoff based on the compute cost/memory cost and quality requirements.
\end{enumerate}

\section*{Future Work}
\label{sec:future_work}

We intend to expand our analysis to additional models, especially those with architectures already optimized for efficiency, such as MobileNet, EfficientNet and Transformers. Furthermore, we would like to extend our research to binary quantization, with the goal to finding even more efficient networks. Another area of future work is to improve training costs with quantization, as this work focuses on optimizing inference costs. In addition, we hope to evaluate our quantized models on real hardware supporting these formats, e.g. NVIDIA A100.
\section*{Conclusion}
\label{sec:conclusion}

In this work, we analyzed how quantization at different precisions influences the compute cost-quality Pareto curves on ResNet models. We found that quantization consistently improves the tradeoff regardless of where we are on the compute cost-quality tradeoff graph. In other words, for each bfloat16 model, we found quantized models with lower compute cost and higher accuracy. Additionally, models in 4-bit (with first and last layers in 8-bit) offer a consistent advantage over 8-bit, using both linear and quadratic cost models. This observation suggests that 4-bit may be a preferred numeric format for quantizing neural networks.
Based on our results, we proposed a simple practical approach to compress models given compute cost and quality constraints, consisting of two steps: quantize the model to 4-bit, then multiply the number of parameters in each layer by a global factor to achieve the desired tradeoff.
We invite further research into comparing different numeric formats for quantization, and hope that this line of work will inform the development of future hardware.
We also encourage more works in model compression to compare methods on cost-quality tradeoff graphs, as this provides a more nuanced and thorough analysis. Lastly, we open-sourced our quantization library, in the hopes that it will accelerate quantization research and deployment with JAX.\\

\textbf{Acknowledgements} We thank Reiner Pope and Mike Gunter for fruitful discussions and feedback on this work. We would also like to thank James Bradbury and Anselm Levskaya for providing invaluable support on JAX/Flax as we implemented our AQT quantization library.

{\small
\bibliographystyle{ieee_fullname}
\bibliography{references}

\begin{thebibliography}{10}\itemsep=-1pt

\bibitem{abadi2016tensorflow}
Mart{\'\i}n Abadi, Paul Barham, Jianmin Chen, Zhifeng Chen, Andy Davis, Jeffrey
  Dean, Matthieu Devin, Sanjay Ghemawat, Geoffrey Irving, Michael Isard, et~al.
\newblock Tensorflow: A system for large-scale machine learning.
\newblock In {\em 12th $\{$USENIX$\}$ symposium on operating systems design and
  implementation ($\{$OSDI$\}$ 16)}, pages 265--283, 2016.

\bibitem{agarwal1995three}
Ramesh~C Agarwal, Susanne~M Balle, Fred~G Gustavson, Mahesh Joshi, and Prasad
  Palkar.
\newblock A three-dimensional approach to parallel matrix multiplication.
\newblock {\em IBM Journal of Research and Development}, 39(5):575--582, 1995.

\bibitem{baskin2018nice}
Chaim Baskin, Natan Liss, Yoav Chai, Evgenii Zheltonozhskii, Eli Schwartz, Raja
  Giryes, Avi Mendelson, and Alexander~M Bronstein.
\newblock Nice: Noise injection and clamping estimation for neural network
  quantization.
\newblock {\em arXiv preprint arXiv:1810.00162}, 2018.

\bibitem{jax2018github}
James Bradbury, Roy Frostig, Peter Hawkins, Matthew~James Johnson, Chris Leary,
  Dougal Maclaurin, George Necula, Adam Paszke, Jake Vander{P}las, Skye
  Wanderman-{M}ilne, and Qiao Zhang.
\newblock {JAX}: composable transformations of {P}ython+{N}um{P}y programs.
\newblock \url{http://github.com/google/jax}, 2018.

\bibitem{cai2020zeroq}
Yaohui Cai, Zhewei Yao, Zhen Dong, Amir Gholami, Michael~W Mahoney, and Kurt
  Keutzer.
\newblock Zeroq: A novel zero shot quantization framework.
\newblock In {\em Proceedings of the IEEE/CVF Conference on Computer Vision and
  Pattern Recognition}, pages 13169--13178, 2020.

\bibitem{choi2018pact}
Jungwook Choi, Zhuo Wang, Swagath Venkataramani, Pierce I-Jen Chuang,
  Vijayalakshmi Srinivasan, and Kailash Gopalakrishnan.
\newblock Pact: Parameterized clipping activation for quantized neural
  networks.
\newblock {\em arXiv preprint arXiv:1805.06085}, 2018.

\bibitem{dai2021vs}
Steve Dai, Rangharajan Venkatesan, Haoxing Ren, Brian Zimmer, William~J Dally,
  and Brucek Khailany.
\newblock Vs-quant: Per-vector scaled quantization for accurate low-precision
  neural network inference.
\newblock {\em arXiv preprint arXiv:2102.04503}, 2021.

\bibitem{dong2019hawq}
Zhen Dong, Zhewei Yao, Amir Gholami, Michael~W Mahoney, and Kurt Keutzer.
\newblock Hawq: Hessian aware quantization of neural networks with
  mixed-precision.
\newblock In {\em Proceedings of the IEEE/CVF International Conference on
  Computer Vision}, pages 293--302, 2019.

\bibitem{elsken2019neural}
Thomas Elsken, Jan~Hendrik Metzen, Frank Hutter, et~al.
\newblock Neural architecture search: A survey.
\newblock {\em J. Mach. Learn. Res.}, 20(55):1--21, 2019.

\bibitem{fang2020post}
Jun Fang, Ali Shafiee, Hamzah Abdel-Aziz, David Thorsley, Georgios Georgiadis,
  and Joseph~H Hassoun.
\newblock Post-training piecewise linear quantization for deep neural networks.
\newblock In {\em European Conference on Computer Vision}, pages 69--86.
  Springer, 2020.

\bibitem{garg2021confounding}
Sahaj Garg, Anirudh Jain, Joe Lou, and Mitchell Nahmias.
\newblock Confounding tradeoffs for neural network quantization.
\newblock {\em arXiv preprint arXiv:2102.06366}, 2021.

\bibitem{han2016deep}
Song Han, Huizi Mao, and William~J. Dally.
\newblock Deep compression: Compressing deep neural networks with pruning,
  trained quantization and huffman coding, 2016.

\bibitem{ResNet}
Kaiming He, Xiangyu Zhang, Shaoqing Ren, and Jian Sun.
\newblock Deep residual learning for image recognition.
\newblock In {\em Proceedings of the IEEE conference on computer vision and
  pattern recognition}, pages 770--778, 2016.

\bibitem{flax2020github}
Jonathan Heek, Anselm Levskaya, Avital Oliver, Marvin Ritter, Bertrand
  Rondepierre, Andreas Steiner, and Marc van {Z}ee.
\newblock {F}lax: A neural network library and ecosystem for {JAX}.
\newblock \url{http://github.com/google/flax}, 2020.

\bibitem{howard2017mobilenets}
Andrew~G Howard, Menglong Zhu, Bo Chen, Dmitry Kalenichenko, Weijun Wang,
  Tobias Weyand, Marco Andreetto, and Hartwig Adam.
\newblock Mobilenets: Efficient convolutional neural networks for mobile vision
  applications.
\newblock {\em arXiv preprint arXiv:1704.04861}, 2017.

\bibitem{jacob2018quantization}
Benoit Jacob, Skirmantas Kligys, Bo Chen, Menglong Zhu, Matthew Tang, Andrew
  Howard, Hartwig Adam, and Dmitry Kalenichenko.
\newblock Quantization and training of neural networks for efficient
  integer-arithmetic-only inference.
\newblock In {\em Proceedings of the IEEE Conference on Computer Vision and
  Pattern Recognition}, pages 2704--2713, 2018.

\bibitem{jain2019trained}
Sambhav~R Jain, Albert Gural, Michael Wu, and Chris~H Dick.
\newblock Trained quantization thresholds for accurate and efficient
  fixed-point inference of deep neural networks.
\newblock {\em arXiv preprint arXiv:1903.08066}, 2019.

\bibitem{kim2020spatial}
Eunhui Kim and Kyong-Ha Lee.
\newblock Spatial shift point-wise quantization.
\newblock {\em IEEE Access}, 8:207683--207690, 2020.

\bibitem{Krizhevsky10convolutionaldeep}
Alex Krizhevsky.
\newblock Convolutional deep belief networks on cifar-10, 2010.

\bibitem{li2018auto}
Guangli Li, Lei Liu, Xueying Wang, Xiao Dong, Peng Zhao, and Xiaobing Feng.
\newblock Auto-tuning neural network quantization framework for collaborative
  inference between the cloud and edge.
\newblock In {\em International Conference on Artificial Neural Networks},
  pages 402--411. Springer, 2018.

\bibitem{mckinstry2018discovering}
Jeffrey~L McKinstry, Steven~K Esser, Rathinakumar Appuswamy, Deepika Bablani,
  John~V Arthur, Izzet~B Yildiz, and Dharmendra~S Modha.
\newblock Discovering low-precision networks close to full-precision networks
  for efficient embedded inference.
\newblock {\em arXiv preprint arXiv:1809.04191}, 2018.

\bibitem{A100_whitepaper}
NVIDIA.
\newblock Nvidia a100 tensor core gpu architecture.
\newblock
  \url{https://www.nvidia.com/content/dam/en-zz/Solutions/Data-Center/nvidia-ampere-architecture-whitepaper.pdf},
  2020.
\newblock Accessed 24 February 2021.

\bibitem{oh2020weight}
Jihun Oh, SangJeong Lee, Meejeong Park, Pooni Walagaurav, and Kiseok Kwon.
\newblock Weight equalizing shift scaler-coupled post-training quantization.
\newblock {\em arXiv preprint arXiv:2008.05767}, 2020.

\bibitem{Pytorch}
Adam Paszke, Sam Gross, Soumith Chintala, Gregory Chanan, Edward Yang, Zachary
  DeVito, Zeming Lin, Alban Desmaison, Luca Antiga, and Adam Lerer.
\newblock Automatic differentiation in pytorch.
\newblock In {\em Conference on Neural Information Processing Systems
  (NeurIPS)}, 2017.

\bibitem{AmoebaNetA}
Esteban Real, Alok Aggarwal, Yanping Huang, and Quoc~V Le.
\newblock Regularized evolution for image classifier architecture search.
\newblock In {\em Proceedings of the aaai conference on artificial
  intelligence}, volume~33, pages 4780--4789, 2019.

\bibitem{ImageNet}
Olga Russakovsky, Jia Deng, Hao Su, Jonathan Krause, Sanjeev Satheesh, Sean Ma,
  Zhiheng Huang, Andrej Karpathy, Aditya Khosla, Michael Bernstein, et~al.
\newblock Imagenet large scale visual recognition challenge.
\newblock {\em International journal of computer vision}, 115(3):211--252,
  2015.

\bibitem{shkolnik2020robust}
Moran Shkolnik, Brian Chmiel, Ron Banner, Gil Shomron, Yuri Nahshan, Alex
  Bronstein, and Uri Weiser.
\newblock Robust quantization: One model to rule them all.
\newblock {\em arXiv preprint arXiv:2002.07686}, 2020.

\bibitem{sun2021effective}
Qigong Sun, Licheng Jiao, Yan Ren, Xiufang Li, Fanhua Shang, and Fang Liu.
\newblock Effective and fast: A novel sequential single path search for
  mixed-precision quantization.
\newblock {\em arXiv preprint arXiv:2103.02904}, 2021.

\bibitem{InceptionV4}
Christian Szegedy, Sergey Ioffe, Vincent Vanhoucke, and Alexander Alemi.
\newblock Inception-v4, inception-resnet and the impact of residual connections
  on learning.
\newblock In {\em Proceedings of the AAAI Conference on Artificial
  Intelligence}, volume~31, 2017.

\bibitem{tan2019efficientnet}
Mingxing Tan and Quoc Le.
\newblock Efficientnet: Rethinking model scaling for convolutional neural
  networks.
\newblock In {\em International Conference on Machine Learning}, pages
  6105--6114. PMLR, 2019.

\bibitem{wang2019haq}
Kuan Wang, Zhijian Liu, Yujun Lin, Ji Lin, and Song Han.
\newblock Haq: Hardware-aware automated quantization with mixed precision.
\newblock In {\em Proceedings of the IEEE/CVF Conference on Computer Vision and
  Pattern Recognition}, pages 8612--8620, 2019.

\bibitem{wu2020integer}
Hao Wu, Patrick Judd, Xiaojie Zhang, Mikhail Isaev, and Paulius Micikevicius.
\newblock Integer quantization for deep learning inference: Principles and
  empirical evaluation.
\newblock {\em arXiv preprint arXiv:2004.09602}, 2020.

\bibitem{Xilinx_whitepaper}
Xilinx.
\newblock Convolutional neural network with int4 optimization on xilinx
  devices.
\newblock
  \url{https://www.xilinx.com/support/documentation/white_papers/wp521-4bit-optimization.pdf},
  2020.
\newblock Accessed 1 March 2021.

\bibitem{zhang2018lq}
Dongqing Zhang, Jiaolong Yang, Dongqiangzi Ye, and Gang Hua.
\newblock Lq-nets: Learned quantization for highly accurate and compact deep
  neural networks.
\newblock In {\em Proceedings of the European conference on computer vision
  (ECCV)}, pages 365--382, 2018.

\bibitem{zhang2021fracbnn}
Yichi Zhang, Junhao Pan, Xinheng Liu, Hongzheng Chen, Deming Chen, and Zhiru
  Zhang.
\newblock Fracbnn: Accurate and fpga-efficient binary neural networks with
  fractional activations.
\newblock In {\em The 2021 ACM/SIGDA International Symposium on
  Field-Programmable Gate Arrays}, pages 171--182, 2021.

\bibitem{zhao2019improving}
Ritchie Zhao, Yuwei Hu, Jordan Dotzel, Chris De~Sa, and Zhiru Zhang.
\newblock Improving neural network quantization without retraining using
  outlier channel splitting.
\newblock In {\em International conference on machine learning}, pages
  7543--7552. PMLR, 2019.

\bibitem{zhuang2018towards}
Bohan Zhuang, Chunhua Shen, Mingkui Tan, Lingqiao Liu, and Ian Reid.
\newblock Towards effective low-bitwidth convolutional neural networks.
\newblock In {\em Proceedings of the IEEE conference on computer vision and
  pattern recognition}, pages 7920--7928, 2018.

\bibitem{zoph2016neural}
Barret Zoph and Quoc~V Le.
\newblock Neural architecture search with reinforcement learning.
\newblock {\em arXiv preprint arXiv:1611.01578}, 2016.

\end{thebibliography}
}
\end{document}